\documentclass[sigconf]{acmart}

\title{Mind the Gap: Linguistic Divergence and Adaptation Strategies in Human-LLM Assistant vs. Human-Human Interactions}

\author{Fulei Zhang}
\affiliation{%
  \institution{Amazon.com Inc.}
  \country{United States}
}
\email{zhanfule@amazon.com}
\author{Zhou Yu}
\affiliation{%
  \institution{Amazon.com Inc.}
  \country{United States}
}
\email{amznzya@amazon.com}

\setcopyright{rightsretained}
\copyrightyear{2025}
\acmYear{2025}
\acmConference[Genaiecom '25]{Proceedings of the second workshop on Generative AI for E-Commerce}{September 22, 2025}{Prague, CZ}
\acmBooktitle{Proceedings of the second workshop on Generative AI for E-Commerce 2025, September 22, 2025}
\acmYear{2025}
\acmMonth{09}
\acmDOI{}
\acmISBN{}



\usepackage[utf8]{inputenc}
\usepackage[T1]{fontenc}
\usepackage{booktabs}
\usepackage{amsmath}
\usepackage{amsfonts}
\usepackage{listings}
\usepackage{algorithm}
\usepackage{algpseudocode}
\usepackage[framemethod=TikZ]{mdframed}
\usepackage{xcolor}
\usepackage{makecell}

\begin{document}

\begin{abstract}
As Large Language Models (LLMs) are increasingly deployed in customer-facing applications, a critical yet underexplored question is how users communicate differently with LLM chatbots compared to human agent. In this study, we present empirical evidence that users adopt distinct communication styles when users interact with chatbots versus human agents. Our analysis reveals significant differences in grammatical fluency, politeness, and lexical diversity in user language between the two settings. These findings suggest that models trained exclusively on human-human interaction data may not adequately accommodate the communication style shift that occurs once an LLM chatbot is deployed. To enhance LLM robustness to post-launch communication style changes, we experimented with two strategies: (1) data augmentation during the post-training phase and (2) inference-time user message reformulation. Our results indicate that models trained on stylistically diverse datasets significantly outperform those trained exclusively on original or stylistically uniform datasets, while inference-time reformulation proved less effective. These insights help us to better adapt our models for improved LLM-user interaction experiences.
\end{abstract}

\maketitle

\section{Introduction}
\label{sec:intro}
Large Language Models (LLMs) offer considerable promise for task-oriented dialogue systems, demonstrating strong capabilities in intent understanding, context retention, commonsense reasoning, and the generation of human-like responses that enhance user experience. In industry conversational chatbot applications, LLM-powered assistants are typically developed and evaluated using historical human-to-human chat transcripts. However, one foundational question often goes unexamined: Do users communicate in the same way with LLM chatbot as they do with human agents?

According to Communication Accommodation Theory (CAT), people naturally adjust their communication style to match or mirror their conversation partners during interactions\cite{dragojevic2015communication}. While LLMs are capable of producing fluent responses, their perceived non-human identity and stylistic tendencies may prompt users to adopt a different linguistic style. As a result, user messages in human-LLM interactions may diverge from those in human-human settings - potentially affecting system performance, particularly when models are trained predominantly on human-human data.

This study addresses this gap. We analyze user messages during the intent understanding phase in multi-turn conversations involving both human agents and LLM-powered chatbots. Our analysis quantitatively demonstrates that people interacting with LLMs use messages that are more terse, grammatically degraded, and less polite compared to those sent to human agents, although these stylistic shifts do not compromise the messages' informativeness or emotional clarity. Such variation introduces a domain/style mismatch and distribution shift in language, creating challenges for models trained primarily on human-to-human interaction data. These models may under-perform when deployed in human-LLM conversations where user communication style changes - not due to task complexity or semantic ambiguity, but because of surface-level linguistic divergences. To overcome this challenge, we experimented with two approaches to improve LLM robustness: interventions at the post-training stage and at inference time. By incorporating stylistically diverse datasets during training, we achieved substantial improvement in intent detection performance on human-LLM assistant interaction messages. In contrast, inference-time user message reformulation did not yield significant performance gains, highlighting the importance of addressing stylistic variations during the training process.

In this work, our contributions are: 
\begin{enumerate}
\item Quantifying linguistic variation in user utterances across human-LLM and human-human interactions using a rubric of six stylistic and semantic dimensions (section ~\ref{sec:cusotmer_lang}).
\item Exploring style-aware data augmentation by generating synthetic user queries spanning different linguistic styles—from ungrammatical and terse to formal and expressive (section ~\ref{sec:cusotmer_lang}). 
\item Demonstrating that stylistically diverse training data significantly boosts performance on real-world human-LLM assistant interaction inputs, outperforming models trained on stylistically consistent datasets (section ~\ref{sec:posttrain}). 
\item Showing that inference-time style normalization under-performs, emphasizing the importance of training-time exposure to natural stylistic diversity(section ~\ref{sec:query}). 
\end{enumerate} 

To our knowledge, there is a lack of empirical studies published explicitly examining users' linguistic adaptation to conversational LLM and to propose approaches for improving LLM robustness to these stylistic variations. Such longitudinal insights could inform the development of more adaptive and context-sensitive conversational LLMs.

\section{Related Work}
Despite the growing popularity of Large Language Models (LLMs) and their promising potential in task-oriented chatbot services such as travel planning, customer support and sales, there remains a significant gap in understanding how users modify their communication patterns when interacting with LLM assistants versus humans. \cite{10.1287/isre.2022.0152} conducted a pioneering study examining how user interactions change when the involvement of human agents in hybrid chat services is either disclosed or concealed. Their findings reveal that users adopt a more human-oriented communication style when aware of human involvement, primarily motivated by impression management concerns. This altered communication style manifests in three key linguistic dimensions: verbosity (using more words per message), complexity (employing more sophisticated sentence structures), and density (higher ratio of function words to content words). However, it is important to note that since the study focused on hybrid systems rather than pure LLM-driven interactions, these findings may not fully represent the communication dynamics in purely LLM guided chat experiences. 

Although no studies have specifically investigated how differences in communication styles between human-LLM assistant and human-human interactions may impact LLM functionality, research on LLM resilience to various input noise types presents conflicting results. Studies on LLM’s resilience on different types of noise such as ASR errors, grammatical /spelling mistakes  and other text irregularities reported different outcome of strong robustness \cite{singh2024robustnessllmsperturbationstext} to suffered performance \cite{wang2024resilience}, possibly due to different noise introduction methods/tasks/types of LLMs. To better handle noisy /diverse text inputs,  post-training techniques\cite{meng2021coco} employ auxiliary tasks and data augmentation strategies to teach models to recognize semantic equivalence despite superficial variations, thereby enhancing the input robustness. Another line of work focuses on improve LLM’s robustness at inference time, either by transforming noisy text inputs to clean input through reference to similar examples\cite{zhang2024noise} or by self-mitigate/correction approaches \cite{jiang2024enhancing}. 

\section{User Language Divergence}
\label{sec:cusotmer_lang}
\subsection{Human-LLM assistant vs. Human-Human Interactions}
To understand how users linguistically adapt depending on their conversational partner (LLM assistant vs. human), we defined six interpretable linguistic dimensions: grammar fluency, politeness/formality, lexical diversity, informativeness, explicitness/clarity, and emotional intensity. We define the linguistic dimensions as follows:

\begin{itemize}
    \item \textbf{Grammar Fluency:} Are the grammar and sentence structure fluent and correct?
    \item \textbf{Politeness/Formality:} Is the tone polite or formal (e.g., ``please,'' ``thank you'') vs.\ informal or blunt?
    \item \textbf{Lexical Diversity:} Does the user use varied and rich vocabulary, as opposed to repetitive or basic wording?
    \item \textbf{Informativeness:} Does the message provide actionable or detailed information relevant to resolving the issue?
    \item \textbf{Explicitness/Clarity:} Is the request clearly stated or vague/ambiguous?
    \item \textbf{Emotion Intensity:} How strongly is the user’s emotion expressed (e.g., frustration, urgency)?
\end{itemize}

Each user message was evaluated on a 1–5 scale using the Claude 3.5 Sonnet v2 model, guided by chain-of-thought prompting to ensure consistent and calibrated judgments across utterances. The full scoring prompt used for this rubric-based evaluation is provided in Appendix~\ref{sec:Language Scoring Prompt}.
\footnote{The dataset and source code cannot be released due to data privacy. We report only relative changes (delta) in metrics.}

Our analysis revealed statistically significant linguistic differences based on the type of conversational partner, as shown in table~\ref{tab:linguistic-dimensions}. user messages directed toward human agents exhibited significantly higher grammar fluency (+5.3\%, $p < 0.001$), greater politeness and formality (+14.5\%, $p < 0.001$), and slightly richer lexical diversity (+1.4\%, $p < 0.05$) compared to those sent to LLM assistants. In contrast, no statistically significant differences were observed in informativeness, explicitness/clarity, or emotional intensity. These results suggest that while users adjust their linguistic style based on the nature of the recipient---being more formal, polite, and grammatically complete with humans---they maintain consistent levels of substantive detail and emotional expression across both interaction types.

These findings suggest that users adapt their linguistic style in human-LLM conversations, producing messages that are shorter, more direct, less formal, and grammatically simpler. This behavior is likely shaped by users’ mental models of LLM chatbot as less socially sensitive or less capable of nuanced interpretation. The result is a stylistic domain shift between human-human and human-LLM conversations.

This stylistic divergence introduces a domain shift, models trained exclusively on polished human-human data may struggle when deployed in real-world LLM assistant environments. As we discuss next, this can be done either by augmenting the training dataset to reflect stylistic variety, or by transforming inference-time queries to better align with the training distribution. 

\begin{table}[th]
  \caption{Comparison of linguistic dimensions between human-LLM and human-human interactions}
  \label{tab:linguistic-dimensions}
  \centering
  \begin{tabular}{l p{4cm} c}
    \toprule
    Linguistic Feature & Relative Difference\newline (Human-Human vs. Human-LLM) & p-value \\
    \midrule
    Grammar Fluency       & +5.3\% & $< 0.001^{**}$ \\
    Politeness/Formality  & +14.5\% & $< 0.001^{**}$ \\
    Lexical Diversity     & +1.4\% & $0.0378^{*}$ \\
    Informativeness       & –0.5\% & 0.535 \\
    Explicitness/Clarity  & –0.5\% & 0.399 \\
    Emotion Intensity     & –0.9\% & 0.258 \\
    \bottomrule
  \end{tabular}
\end{table}

\subsection{Learning from Style-Augmented Datasets}
To address the linguistic style mismatch identified in Section 3.1, we propose a post-training phase mitigation strategy that synthesizes new data reflecting diverse stylistic variants. In our setting, we only utilize conversation data from human to human interactions and evaluate our results on conversations from human to LLM assistant interactions. 

Using controlled prompting with a large language model (Claude 3.5 Sonnet v2), we synthetically rewrite original user messages into two complementary styles: a "Minimal Style" version that exhibits characteristics in interactions with LLM assistant: low grammar fluency, impoliteness, and low lexical diversity, and an "Enriched Style" version that simulates exaggerated human-human communication with high fluency, formality, and vocabulary richness. In both settings, we explicitly instructed the model to preserve the original meaning and avoid introducing new content. 

The rewriting procedure is formalized in Algorithm~\ref{alg:rewriting}. Rewrites were generated by adjusting each message to align with the linguistic scores. Full prompt templates for both the Minimal Style and Enriched Style rewrites are provided in Appendix~\ref{sec:minimal-style} and Appendix~\ref{sec:enriched-style}, respectively.

\begin{algorithm}[h]
\caption{Controlled Rewriting Strategy}
\label{alg:rewriting}
\begin{algorithmic}[1]
\Require User message $u$, original attribute scores $s = (s_g, s_p, s_l)$
\Ensure Rewritten message $u'$
\Comment{$s_g$ = grammar fluency, $s_p$ = politeness/formality, $s_l$ = lexical diversity}
\If{Minimal Style}
    \State $t \gets (\max(1, s_g - 1), \max(1, s_p - 1), \max(1, s_l - 1))$
\ElsIf{Enriched Style}
    \State $t \gets (\min(5, s_g + 1), \min(5, s_p + 1), \min(5, s_l + 1))$
\EndIf
\State Prompt LLM with $(u, s, t)$ using rewrite template
\State $u' \gets$ LLM output
\State \Return $u'$
\end{algorithmic}
\end{algorithm}

To illustrate the stylistic transformations produced by our controlled rewriting strategy, consider the following example:

\begin{mdframed}[backgroundcolor=gray!5,linewidth=0.5pt]
\textbf{Original Message $(s_g{=}3,\, s_p{=}3,\, s_l{=}3)$:} \\
Hi, I’m looking to plan a trip to Paris next month. Can you help me find good flight and hotel options?

\smallskip
\textbf{Minimal Style Rewrite $(s_g{=}1,\, s_p{=}1,\, s_l{=}1)$:} \\
paris next month. flights hotels?

\smallskip
\textbf{Enriched Style Rewrite $(s_g{=}5,\, s_p{=}5,\, s_l{=}5)$:} \\
Good afternoon! I'm planning a vacation to Paris in the coming month and would appreciate your help finding the best deals on both flights and accommodations. Thank you!
\end{mdframed}

This example highlights the lexical, grammatical, and tonal shifts captured by our rewriting prompts. The minimal style reflects the kind of terse, blunt phrasing typical of human-LLM interactions, while the enriched style exaggerates the fluency and formality characteristic of human-human interactions.

\subsection{Reformulate Human-LLM Assistant Inquiry}
An alternative to training-time augmentation is to align user input style during inference, while still rely on models trained only on human-human interactions. This approach, similar to query reformulation, aims to convert LLM interaction-style queries into human interaction-style variants prior to prediction. 

To apply this idea, we evaluate incoming user messages along the grammar, politeness, and lexical diversity dimensions. If a message scores above a threshold in all dimensions (>2), we preserve it. Otherwise, we rewrite it to match a randomly sampled style vector from our human-human dataset using Claude 3.5 Sonnet v2.

This reformulation process aims to make test-time queries more familiar to a model trained exclusively on human-human data. However, as discussed in later sections, this technique may risk altering subtle semantic cues, raising concerns about unintended information loss.

Therefore, this section outlines a practical inference-time adaptation mechanism that complements our training-side strategy in Section 3.2. Together, these techniques offer a comprehensive framework for managing linguistic variability in human-LLM interactions.

\section{Experimental Framework}
\subsection{Task}
We study the task of intent understanding in the context of task-oriented dialogues. This task occurs early in a conversation—typically at the first user turn—before the dialogue is shaped by system policies or human interventions. Given a user’s initial message, the objective is to identify the underlying intent from a predefined list of possible intent classes. This task is important for downstream applications such as automated routing, response generation, and case resolution. 

We formulate this as a generative classification problem, where a large language model (LLM) is instruction-finetuned to produce the appropriate intent label in free-text form. At inference time, the model receives a standardized instruction, a list of candidate intents, and a user message, and is expected to generate the label that best reflects the user’s need. The intent space includes a wide range of service-related categories and requires the model to distinguish between subtle semantic variations in user input.

\subsection{Data}
Our training set comprises 13K user utterances drawn from anonymized human-to-human chat transcripts. To ensure clear intent signals, we extract only the initial user message from each session, avoiding contamination from follow-up clarifications or human guidance. Non-informative utterances such as greetings, pleasantries, or empty inputs are excluded. Each message is annotated with one of intent categories drawn from a standardized user intent ontology.

For evaluation, we use a held-out test set of 1,357 anonymized user messages from real-world conversations with LLM-based chatbots. This test setting reflects a realistic deployment context where models trained on human-human dialogue are applied to human-LLM interactions, often exhibiting stylistic or structural shifts in user behavior.

\subsection{Evaluation Metrics}
We evaluate model performance using accuracy, defined as the proportion of test-set examples where the LLM-generated intent label exactly matches the annotated ground truth label. Since each input corresponds to a single intent from a fixed label set, accuracy is an appropriate and interpretable metric for this setup.

\section{Results}
\subsection{Post-Training with Diverse Linguistic Style}
\label{sec:posttrain}
To evaluate the impact of user linguistic style on user intent prediction task performance, we conducted a series of post-training data augmentation experiments. Our goal was to test whether exposing the model to a broader range of language styles - from grammatically minimal and impolite to fluent and formal, would improve its ability to generalize to real-world human-LLM interaction style messages.

We experiment with four training configurations:
\begin{itemize}
    \item \textbf{$D_1$ (Human-human):} Original user messages with human agent.

    \item \textbf{$D_2$ (Minimal Style):} Rewritten messages with low grammar fluency, low politeness, and low lexical diversity, simulating terse, informal LLM-directed input.

    \item \textbf{$D_3$ (Enriched Style):} Rewritten messages with high scores on grammar, politeness, and vocabulary richness, simulating more formal user interactions.

    \item \textbf{$D_4$ (Combined):} Union of $D_1$, $D_2$, and $D_3$ for maximum style variation.
\end{itemize}

To validate the effectiveness of our controlled rewriting prompts, we re-scored all rewritten samples across grammar fluency, politeness/formality, and lexical diversity using our rubric scoring prompt in Appendix~\ref{sec:Language Scoring Prompt}. As shown in Table~\ref{tab:linguistic-dataset-variants}, this retrospective analysis confirms that the rewrites reliably shifted the linguistic characteristics in the intended directions. Compared to the original human-human dataset ($D_1$), the Minimal Style rewrites ($D_2$) exhibit lower scores across all dimensions, while the Enriched Style rewrites ($D_3$) achieve significantly higher scores. The Combined dataset ($D_4$), which includes all styles, displays a balanced average - capturing the stylistic diversity needed to improve generalization.

\begin{table}[ht]
  \caption{Relative differences in linguistic dimension scores across training dataset variants}
  \label{tab:linguistic-dataset-variants}
  \centering
  \small
  \begin{tabular}{l c c c}
    \toprule
    Dataset & \makecell{Grammar\\Fluency} & \makecell{Politeness/\\Formality} & \makecell{Lexical\\Diversity} \\
    \midrule
    D\textsubscript{1} (Human-human) & 0.0\%   & 0.0\%   & 0.0\%   \\
    D\textsubscript{2} (Minimal Style) & –15.8\% & –18.7\% & –12.8\% \\
    D\textsubscript{3} (Enriched Style)  & +56.9\% & +67.5\% & +47.7\% \\
    D\textsubscript{4} (Combined)& +35.3\% & +44.5\% & +31.1\% \\
    \bottomrule
  \end{tabular}
\end{table}

All models were fine-tuned using LoRA~\cite{hulora} on the Mistral-7B base model~\cite{jiang2024enhancing}, and evaluated on a held-out test set of real-world user–LLM assistant conversations. This setup simulates a realistic deployment context where models trained on human-human transcripts must generalize to user utterances that differ stylistically in human-LLM settings.

As shown in Table~\ref{tab:intent-accuracy-relative}, the model trained on the combined dataset ($D_4$) achieved the best performance, with a \textbf{+2.9\%} relative improvement over the baseline ($D_1$). This suggests that exposure to a range of linguistic styles—including formal, informal, and terse utterances—enhances the model's ability to generalize to real-world chatbot inputs.

By contrast, models trained solely on minimal-style ($D_2$) or enriched-style ($D_3$) data underperformed relative to the baseline, with \textbf{–2.6\%} and \textbf{–1.8\%} drops in accuracy, respectively. Despite $D_2$ stylistically resembling typical language in user-LLM interactions, its narrow style range reduced generalization. These findings highlight the importance of stylistic diversity in training data: intent detection systems perform best when exposed to a broad spectrum of real-world user expression rather than a single linguistic register.

\begin{table}[h]
  \caption{Performance change in intent detection accuracy on human-LLM inputs. All values are computed as delta with respect to the baseline model trained on human-human data ($D_1$).}
  \label{tab:intent-accuracy-relative}
  \centering
  \begin{tabular}{lc}
    \toprule
    \textbf{Training Dataset} & \textbf{$\Delta$ vs.\ $D_1$} \\
    \midrule
    $D_1$: Human-human           & 0.0\% \\
    $D_2$: Minimal Style         & –2.6\% \\
    $D_3$: Enriched Style        & –1.8\% \\
    $D_4$: Combined              & \textbf{+2.9\%}$^{**}$ \\
    \bottomrule
  \end{tabular}
\end{table}

\subsection{Query Reformulation at Inference Time}
\label{sec:query}

To mitigate the domain mismatch caused by stylistic variation, we also explored a lightweight alternative to retraining: query reformulation at inference time. This approach assesses whether aligning user inputs to resemble the style of training data - without modifying the model itself, can improve performance. Practically, this method offers an appealing lightweight solution, as it avoids the need for retraining and can be integrated as a pre-processing module in production pipelines.

For evaluation, we used the model trained solely on human-human data ($D_1$) and applied a controlled rewriting process to test inputs from human-LLM conversations. Each message was scored along grammar fluency, politeness/formality, and lexical diversity dimensions using our rubric-based evaluator. If a message scored above threshold across all dimensions, it was retained unchanged. Otherwise, we sampled a target style vector from the $D_1$ distribution and used \textbf{Claude 3.5 Sonnet v2} to rewrite the message to match the target style while preserving the original meaning and intent.

As shown in Table~\ref{tab:inference-rewrite-relative}, inference-time rewriting resulted in a \textbf{–1.9\% relative drop} in performance compared to the original inputs. This result suggests that simply restyling input text to match the training distribution may fail to preserve subtle intent-relevant signals present in original user messages. In some cases, rewriting may introduce unnatural phrasing or obscure key cues critical for classification.

These findings reinforce the key insight of our study: training-time exposure to diverse linguistic variation is more effective than inference-time normalization. Models must learn to interpret diverse communication styles during training, rather than rely on brittle post-hoc transformations that risk semantic distortion.

\begin{table}[ht]
  \caption{Impact of inference-time query reformulation on intent detection accuracy. Results shown as change compared to original input.}
  \label{tab:inference-rewrite-relative}
  \centering
  \begin{tabular}{l c}
    \toprule
    Inference Input Style & Accuracy Delta \\
    \midrule
    Original human-LLM input & 0.0\% \\
    Rewritten to human-human style & \textbf{–1.9\%} \\
    \bottomrule
  \end{tabular}
\end{table}

\section{Conclusion and Future Work}
Through in-depth analysis of task oriented multi-turn conversations from both user-LLM and user-human interactions, we identified user messages exhibit significant communication style differences in these two settings. We quantified these differences across six dimensions, finding substantial variations in grammatical fluency, politeness/formality, and lexical diversity. To better adapt models initially trained and evaluated on human-human communication data to the observed shifts in user communication style post-deployment, we conducted experiments using an intent detection task, exploring both post-training data augmentation techniques and inference-time message reformulation approaches. Our results demonstrate that increasing stylistic diversity in post-training data significantly improves model performance on user-LLM assistant conversations, while inference-time message reformulation proves less effective. This study provides valuable insights into accommodating users' varied linguistic behaviors when interacting with LLM-based systems, enabling more robust conversational LLM that can deliver optimal user experience. While our work primarily focused on the initial phase of conversation and intent detection tasks, future research should investigate how conversational LLM can maintain engaging interactions throughout extended dialogues.

\bibliographystyle{ACM-Reference-Format}

\begin{thebibliography}{8}


\ifx \showCODEN    \undefined \def \showCODEN     #1{\unskip}     \fi
\ifx \showDOI      \undefined \def \showDOI       #1{#1}\fi
\ifx \showISBNx    \undefined \def \showISBNx     #1{\unskip}     \fi
\ifx \showISBNxiii \undefined \def \showISBNxiii  #1{\unskip}     \fi
\ifx \showISSN     \undefined \def \showISSN      #1{\unskip}     \fi
\ifx \showLCCN     \undefined \def \showLCCN      #1{\unskip}     \fi
\ifx \shownote     \undefined \def \shownote      #1{#1}          \fi
\ifx \showarticletitle \undefined \def \showarticletitle #1{#1}   \fi
\ifx \showURL      \undefined \def \showURL       {\relax}        \fi
\providecommand\bibfield[2]{#2}
\providecommand\bibinfo[2]{#2}
\providecommand\natexlab[1]{#1}
\providecommand\showeprint[2][]{arXiv:#2}

\bibitem[Dragojevic et~al\mbox{.}(2015)]%
        {dragojevic2015communication}
\bibfield{author}{\bibinfo{person}{Marko Dragojevic}, \bibinfo{person}{Jessica Gasiorek}, {and} \bibinfo{person}{Howard Giles}.} \bibinfo{year}{2015}\natexlab{}.
\newblock \showarticletitle{Communication accommodation theory}.
\newblock In \bibinfo{booktitle}{\emph{The International Encyclopedia of Interpersonal Communication}}, \bibfield{editor}{\bibinfo{person}{Charles~R. Berger} {and} \bibinfo{person}{Michael~E. Roloff}} (Eds.). \bibinfo{publisher}{Wiley Blackwell}, \bibinfo{pages}{1--21}.
\newblock
\urldef\tempurl%
\url{https://doi.org/10.1002/9781118540190.wbeic006}
\showDOI{\tempurl}


\bibitem[Gnewuch et~al\mbox{.}(2024)]%
        {10.1287/isre.2022.0152}
\bibfield{author}{\bibinfo{person}{Ulrich Gnewuch}, \bibinfo{person}{Stefan Morana}, \bibinfo{person}{Oliver Hinz}, \bibinfo{person}{Ralf Kellner}, {and} \bibinfo{person}{Alexander Maedche}.} \bibinfo{year}{2024}\natexlab{}.
\newblock \showarticletitle{More Than a Bot? The Impact of Disclosing Human Involvement on Customer Interactions with Hybrid Service Agents}.
\newblock \bibinfo{journal}{\emph{Info. Sys. Research}} \bibinfo{volume}{35}, \bibinfo{number}{3} (\bibinfo{date}{Sept.} \bibinfo{year}{2024}), \bibinfo{pages}{936–955}.
\newblock
\showISSN{1526-5536}
\urldef\tempurl%
\url{https://doi.org/10.1287/isre.2022.0152}
\showDOI{\tempurl}


\bibitem[Hu et~al\mbox{.}(2022)]%
        {hulora}
\bibfield{author}{\bibinfo{person}{Edward~J Hu}, \bibinfo{person}{Phillip Wallis}, \bibinfo{person}{Zeyuan Allen-Zhu}, \bibinfo{person}{Yuanzhi Li}, \bibinfo{person}{Shean Wang}, \bibinfo{person}{Lu Wang}, \bibinfo{person}{Weizhu Chen}, {et~al\mbox{.}}} \bibinfo{year}{2022}\natexlab{}.
\newblock \showarticletitle{LoRA: Low-Rank Adaptation of Large Language Models}. In \bibinfo{booktitle}{\emph{International Conference on Learning Representations}}.
\newblock


\bibitem[Jiang et~al\mbox{.}(2024)]%
        {jiang2024enhancing}
\bibfield{author}{\bibinfo{person}{Ming Jiang}, \bibinfo{person}{Tingting Huang}, \bibinfo{person}{Biao Guo}, \bibinfo{person}{Yao Lu}, {and} \bibinfo{person}{Feng Zhang}.} \bibinfo{year}{2024}\natexlab{}.
\newblock \showarticletitle{Enhancing robustness in large language models: Prompting for mitigating the impact of irrelevant information}.
\newblock \bibinfo{journal}{\emph{arXiv preprint arXiv:2408.10615}} (\bibinfo{year}{2024}).
\newblock


\bibitem[Meng et~al\mbox{.}(2021)]%
        {meng2021coco}
\bibfield{author}{\bibinfo{person}{Yu Meng}, \bibinfo{person}{Chenyan Xiong}, \bibinfo{person}{Payal Bajaj}, \bibinfo{person}{Paul Bennett}, \bibinfo{person}{Jiawei Han}, \bibinfo{person}{Xia Song}, {et~al\mbox{.}}} \bibinfo{year}{2021}\natexlab{}.
\newblock \showarticletitle{Coco-lm: Correcting and contrasting text sequences for language model pretraining}.
\newblock \bibinfo{journal}{\emph{Advances in Neural Information Processing Systems}}  \bibinfo{volume}{34} (\bibinfo{year}{2021}), \bibinfo{pages}{23102--23114}.
\newblock


\bibitem[Singh et~al\mbox{.}(2024)]%
        {singh2024robustnessllmsperturbationstext}
\bibfield{author}{\bibinfo{person}{Ayush Singh}, \bibinfo{person}{Navpreet Singh}, {and} \bibinfo{person}{Shubham Vatsal}.} \bibinfo{year}{2024}\natexlab{}.
\newblock \bibinfo{title}{Robustness of LLMs to Perturbations in Text}.
\newblock
\newblock
\showeprint[arxiv]{2407.08989}~[cs.CL]
\urldef\tempurl%
\url{https://arxiv.org/abs/2407.08989}
\showURL{%
\tempurl}


\bibitem[Wang et~al\mbox{.}(2024)]%
        {wang2024resilience}
\bibfield{author}{\bibinfo{person}{Bin Wang}, \bibinfo{person}{Chengwei Wei}, \bibinfo{person}{Zhengyuan Liu}, \bibinfo{person}{Geyu Lin}, {and} \bibinfo{person}{Nancy Chen}.} \bibinfo{year}{2024}\natexlab{}.
\newblock \showarticletitle{Resilience of Large Language Models for Noisy Instructions}. In \bibinfo{booktitle}{\emph{Findings of the Association for Computational Linguistics: EMNLP 2024}}. \bibinfo{pages}{11939--11950}.
\newblock


\bibitem[Zhang et~al\mbox{.}(2024)]%
        {zhang2024noise}
\bibfield{author}{\bibinfo{person}{Feipeng Zhang}, \bibinfo{person}{Wenyu Jiang}, \bibinfo{person}{Jun Shu}, \bibinfo{person}{Feng Zheng}, \bibinfo{person}{Hongxin Wei}, {et~al\mbox{.}}} \bibinfo{year}{2024}\natexlab{}.
\newblock \showarticletitle{On the noise robustness of in-context learning for text generation}.
\newblock \bibinfo{journal}{\emph{Advances in Neural Information Processing Systems}}  \bibinfo{volume}{37} (\bibinfo{year}{2024}), \bibinfo{pages}{16569--16600}.
\newblock


\end{thebibliography}

\appendix
\section{Prompt Templates}

\subsection{Language Scoring Prompt}
\label{sec:Language Scoring Prompt}
\begin{lstlisting}[label={lst:rubric-scoring}]
I will give you a single user utterance. Your task is to evaluate the language used by the user. Use a chain-of-thought approach to reason through your judgments and output a structured JSON dictionary of scores.

Each score should be on a scale from 1 to 5, where 1 = very low / poor, and 5 = very high / excellent. For emotion categories, list the most likely one(s).

Evaluate the following dimensions:

1. Linguistic Features
  - GrammarFluency: Are the grammar and sentence structure fluent and correct?
  - PolitenessFormality: Is the tone polite or formal (e.g., "please", "thank you")? Or informal/slangy?
  - LexicalDiversity: Does the user use varied and rich vocabulary?

2. Semantic Features
  - Informativeness: Does the utterance provide actionable or detailed information?
  - ExplicitnessClarity: Is the request clearly stated or vague?

3. Emotional Features 
  - EmotionIntensity: How strongly is the emotion expressed?

Think step-by-step. First, examine the grammar, politeness, and vocabulary. Then evaluate informativeness and clarity. Finally, assess emotional tone and intensity.

Return a JSON object only.

Begin reasoning now for the following utterance:
{{rewritten_text}}
\end{lstlisting}

\subsection{Minimal Style Rewriting Prompt (D\textsubscript{2})}
\label{sec:minimal-style}
\begin{lstlisting}[label={lst:minimal-rewrite}]
You are a user message rewriting assistant. Your task is to rewrite user messages according to three language attributes while preserving the original meaning and informativeness.

Each attribute is rated from 1 (very low/poor) to 5 (very high/excellent):
  1. GrammarFluency: Are the grammar and sentence structure fluent and correct?
  2. PolitenessFormality: Is the tone polite or formal (e.g., "please", "thank you")? Or informal/slangy?
  3. LexicalDiversity: Does the user use varied and rich vocabulary?

If the rewrite action is REWRITE:
  - Rewrite the message to reflect the target scores, especially when scores are low (e.g., 1 or 2).
  - Lower GrammarFluency = broken, fragmented, ungrammatical sentence.
  - Lower PolitenessFormality = no "please", "thanks", or polite phrasing.
  - Lower LexicalDiversity = repetitive, simple, blunt words.
  - The rewrite should be short, direct, and minimal if target scores are low.
  - Do not add or infer anything not in the original message.

If the rewrite action is KEEP:
  - Return the original message unchanged.

Output only the rewritten message. Do not explain or include any prefix or reasoning.

Original Message: {{processed_turn_text}}
Original Scores: GrammarFluency: {{grammar_fluency}}, PolitenessFormality: {{politeness_formality}}, LexicalDiversity: {{lexical_diversity}}
Target Scores: GrammarFluency: {{target_grammar_fluency}}, PolitenessFormality: {{target_politeness_formality}}, LexicalDiversity: {{target_lexical_diversity}}
Rewrite Action: {{rewrite_action}}
\end{lstlisting}

\subsection{Enriched Style Rewriting Prompt (D\textsubscript{3})}
\label{sec:enriched-style}
\begin{lstlisting}[label={lst:enriched-rewrite}]
You are a user message improvement assistant. Your task is to rewrite user messages to improve their language across three attributes, while keeping the original meaning and intent unchanged.

Each attribute is rated from 1 (very low/poor) to 5 (very high/excellent):
GrammarFluency: Use fluent, grammatically correct, and complete sentence structures.
PolitenessFormality: Use polite or formal tone (e.g., "please", "thank you", "could you"), where appropriate.
LexicalDiversity: Use varied, expressive, and natural vocabulary.

When target scores are high (4 or 5), your goal is to:
- Improve sentence structure to be clear and fluent.
- Add softeners and polite language.
- Use more varied and natural vocabulary while preserving the original meaning.

Do not change the user's intent, add extra information, or make the message longer than necessary.

Only return the rewritten message. Do not explain your reasoning or include commentary.

Original Message: {{processed_turn_text}}
Original Scores: GrammarFluency: {{grammar_fluency}}, PolitenessFormality: {{politeness_formality}}, LexicalDiversity: {{lexical_diversity}}
Target Scores: GrammarFluency: {{target_grammar_fluency}}, PolitenessFormality: {{target_politeness_formality}}, LexicalDiversity: {{target_lexical_diversity}}
Rewrite Action: REWRITE
\end{lstlisting}

\end{document}